\documentclass[10pt,twocolumn,letterpaper]{article}

\usepackage[pagenumbers]{cvpr} 

\definecolor{cvprblue}{rgb}{0.21,0.49,0.74}
\usepackage[pagebackref,breaklinks,colorlinks,allcolors=cvprblue]{hyperref}
\usepackage{url}
\usepackage{graphicx}
\usepackage{hyperref}
\usepackage{multirow}
\usepackage{verbatim}
\usepackage{pifont}
\usepackage{xcolor}
\graphicspath{{./fig/}}

\newcommand{\brown}[1]{\textcolor{brown}{#1}}

\def\paperID{371} 
\def\confName{CVPR}
\def\confYear{2025}

\title{AnyText2: Visual Text Generation and Editing With Customizable Attributes}

\author{Yuxiang Tuo, Yifeng Geng, Liefeng Bo \\
Tongyi Lab, Alibaba Group\\
\texttt{\{yuxiang.tyx,cangyu.gyf,liefeng.bo\}}  \\
\texttt{@alibaba-inc.com}}

\begin{document}
\maketitle
\begin{abstract}

As the text-to-image (T2I) domain progresses, generating text that seamlessly integrates with visual content has garnered significant attention. However, even with accurate text generation, the inability to control font and color can greatly limit certain applications, and this issue remains insufficiently addressed. This paper introduces \textbf{AnyText2}, a novel method that enables precise control over multilingual text attributes in natural scene image generation and editing. Our approach consists of two main components. First, we propose a WriteNet+AttnX architecture that injects text rendering capabilities into a pre-trained T2I model. Compared to its predecessor, AnyText, our new approach not only enhances image realism but also achieves a 19.8\% increase in inference speed. Second, we explore techniques for extracting fonts and colors from scene images and develop a Text Embedding Module that encodes these text attributes separately as conditions. As an extension of AnyText, this method allows for customization of attributes for each line of text, leading to improvements of 3.3\% and 9.3\% in text accuracy for Chinese and English, respectively. Through comprehensive experiments, we demonstrate the state-of-the-art performance of our method. The code and model will be made open-source in \url{https://github.com/tyxsspa/AnyText2}.

\end{abstract}    
\section{Introduction}

\vspace{0mm}
\begin{figure*}[htbp]
 \centering
 \includegraphics[width=1.0\textwidth]{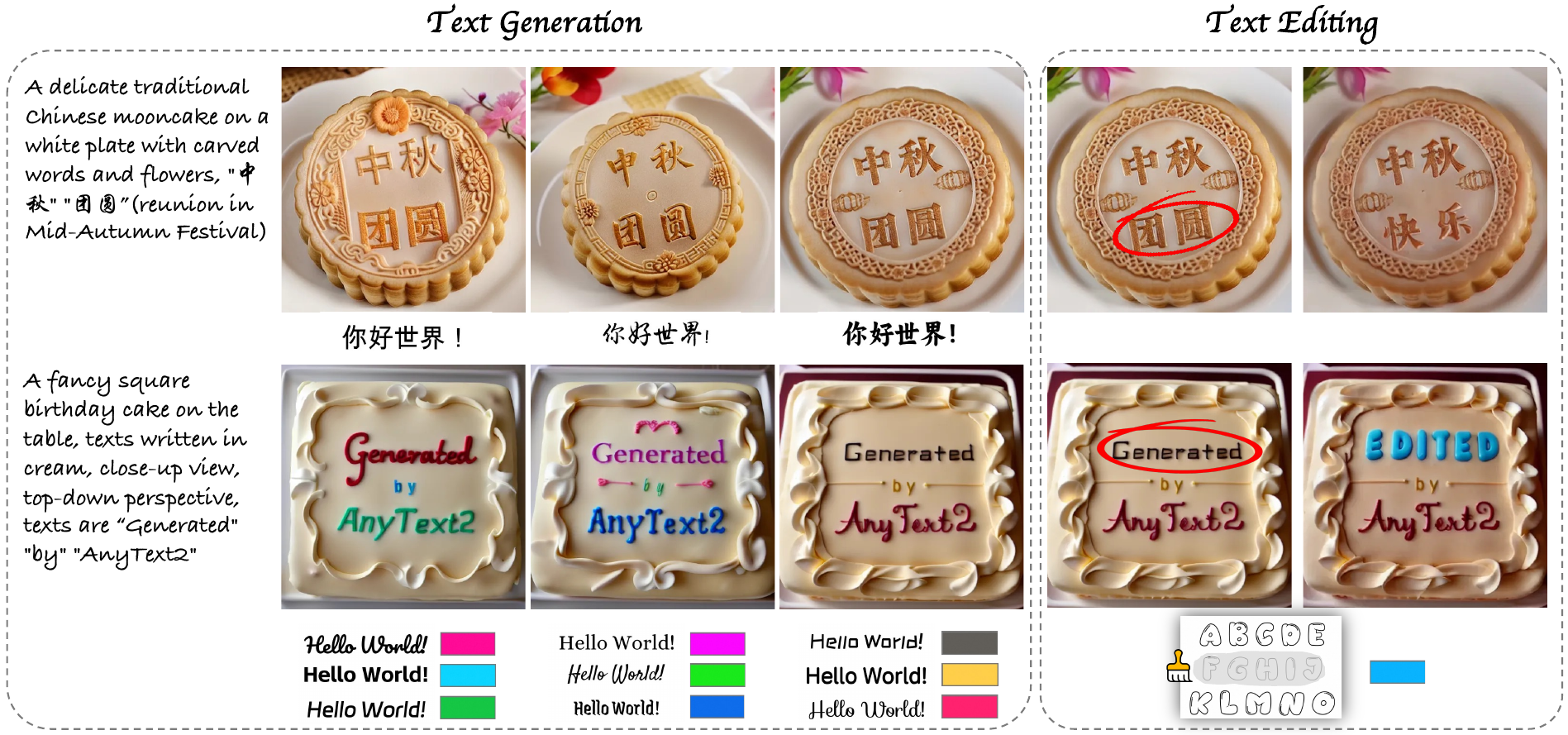}
 \caption{AnyText2 can accurately generate multilingual text within images and achieve a realistic integration. Furthermore, it allows for customized attributes for each line, such as controlling the font style through font files, mimicking an image using a brush tool, and specifying the text color. Additionally, AnyText2 enables customizable attribute editing of text within images.}
 \vspace{-5mm}
 \label{fig:samples}
\end{figure*}

Diffusion-based generative models~\cite{Ho_DDPM_NIPS20, Rombach_LDM_CVPR22, Ramesh_DALLE_icml21, Ramesh_DALLE2_corr22, Dustin_SDXL_ICLR24} have gained prominence due to their ability to generate highly realistic images with intricate details, and gradually replacing previous technologies like GANs~\cite{Goodfellow_GAN_NPIS14} and VAEs~\cite{Diederik_VAE_ICLR14}. In recent research, models such as DALL·E3~\cite{DALLE3}, Stable Diffusion 3~\cite{Patrick_SD3_ICML24}, and FLUX.1~\cite{FLUX.1} have enhanced their visual text rendering capabilities through the introduction of new technologies, such as encoding image captions using large language models like T5, or employing rectified flow transformers. However, the performance of these general-purpose T2I models in text rendering still falls short of expectations. Therefore, many researchers aim to inject enhanced text rendering capabilities into pre-trained diffusion models using various technical methods while maintaining their diversity and realism in image synthesis. These methods, such as GlyphDraw~\cite{Ma_glyphdraw_Corr23}, GlyphControl~\cite{Yang_GlyphControl_Corr23}, TextDiffuser~
\cite{Chen_TextDiffuser_Corr23}, AnyText~\cite{tuo2023anytext}, TextDiffuser-2~\cite{Chen_TextDiffuser2}, Glyph-SDXL~\cite{glyphbyt5}, Glyph-SDXL-v2~\cite{glyphbyt5-v2}, GlyphDraw2~\cite{glyphdraw2}, not only significantly improves the accuracy of text rendering but also extends functionalities such as multilingual text generation, text editing, automatic or specified layout, and even customizable text attributes.

There are two primary mechanisms for injecting text rendering capabilities into pre-trained models: (1) \textit{conditional embeddings} in the prompt and (2) \textit{auxiliary pixels} in the latent space. The first approach encodes the visual appearance of each character as embeddings and combines them with image captions, as seen in methods like TextDiffuser-2, Glyph-SDXL, and Glyph-SDXL-v2. The second approach utilizes text layout masks, injecting them into the latent space as catalysts for text rendering, exemplified by methods such as TextDiffuser and GlyphControl. However, both methods have drawbacks: conditional embeddings struggle to generalize to unseen characters, while auxiliary pixels suffer from limited integration due to a lack of fusion between pre-rendered characters and image captions. To address these limitations, methods like AnyText and GlyphDraw2 employ a combined strategy. Our proposed method, AnyText2, also adopts this mechanism but diverges by utilizing the WriteNet+AttnX architecture. This architecture decouples text rendering from image generation before performing text-image fusion, resulting in a streamlined approach that significantly increases inference speed while improving image realism.

In practical applications, users often require not only legible text embedded in images but also the ability to specify attributes such as font and color. This capability is particularly important for generating or editing designs for logos, advertisement posters, and product showcase images. General-purpose T2I models are unable to achieve effective control over fonts and colors described textually in captions. Recent works, such as DiffSTE~\cite{diffste} and Glyph-SDXL~\cite{glyphbyt5}, have attempted to address this issue by training on synthetic image data, significantly enhancing the understanding of font and color descriptions in captions. However, this approach has two drawbacks: first, any font or color names not included in the training dataset will not be recognized; second, training on synthetic data primarily produces \textit{overlaid} text rather than \textit{embedded} text, which limits its applicability in real-world contexts. To overcome these limitations, we propose a novel method that extracts text attribute labels from natural scene images and introduces a Text Embedding Module that encodes each attribute as individual conditions. This offers great flexibility during the application, as users no longer need to describe text attributes in captions; instead, they can control fonts by specifying font files or by uploading images containing text in a particular style. Color can be controlled through a color picker or palette. More importantly, our method generates both overlaid and embedded text applicable in open-domain scenarios. Selected examples are presented in Fig.~\ref{fig:samples}.
\section{Related Works}
\vspace{-0.1cm}
\noindent \textbf{Controllable Text-To-Image Generation}  In T2I models, achieving precise control through pure textual descriptions poses significant challenges, and a multitude of methods have emerged. Among the pioneering works are ControlNet~\cite{Zhang_ControlNet_Corr23}, T2I Adapter~\cite{Mou_T2I_Corr23}, and Composer~\cite{Huang_Composer_Corr23},  leverage control conditions such as depth maps, pose images, and sketches to guide image generation. Another category is comprised of subject-driven methods, such as Textual Inversion~\cite{Gal_Inversion_ICLR23}, DreamBooth~\cite{Ruiz_DreamBooth_Corr22}, IP-Adapter~\cite{ip-adapter}, ReferenceNet~\cite{hu_animateanyone}, InstantID~\cite{wang2024_instantid}, and PhotoMaker~\cite{li2023photomaker}. These methods focus on learning the representation of a specific subject or concept from one or a few images, primarily ensuring identity preservation in the generated images while allowing less stringent control over other attributes such as position, size, and orientation. Visual text generation can be viewed as a sub-task within this framework if we consider each character as an identity.

\noindent \textbf{Visual Text Generation} The text encoder plays a crucial role in generating accurate visual text, as highlighted by ~\cite{Liu_CharacterAware_ACL23}. Many subsequent methods adopted character-level text encoders to incorporate word spelling or character visual appearance into conditional embeddings, such as DiffSTE~\cite{diffste}, TextDiffuser-2~\cite{Chen_TextDiffuser2}, UDiffText~\cite{zhao2023udifftext}, and Glyph-SDXL~\cite{glyphbyt5}.  In contrast, TextDiffuser~\cite{Chen_TextDiffuser_Corr23} and GlyphControl~\cite{Yang_GlyphControl_Corr23} utilize text layout masks and inject them into the latent space to assist in text generation.  Furthermore, AnyText~\cite{tuo2023anytext} and GlyphDraw2~\cite{glyphdraw2} leverage pre-rendered glyph images as auxiliary pixels and employ a pre-trained OCR model~\cite{chen_ppocrv3_corr22} to encode strokes, which are then integrated into the conditional embeddings. Additionally, methods like UDiffText, Brush Your Text~\cite{zhang2023_brushyourtext}, and Glyph-SDXL impose restrictive interventions on attention maps in corresponding text areas to improve accuracy. In contrast, our approach employs a position encoder to encode text position and injects it into the conditional embeddings, allowing for spatial awareness.

\noindent \textbf{Text Attributes Customization} Numerous studies have explored font style transfer using GANs or diffusion models, categorized as Few-shot Font Generation (FFG), including LF-Font~\cite{Park_LFFont_AAA21}, MX-Font~\cite{Park_MXFont_ICCV21}, Diff-Font~\cite{He_DiffFont_IJCV24}, and FontDiffuser~\cite{yang2024fontdiffuser}. While our work shares the goal of decoupling content and style from reference characters, we focus on generating text in a specified style directly onto an image, as opposed to automatically creating a library of plain characters. Regarding color control, general-purpose T2I models often have difficulty interpreting RGB values specified in prompts. To address this issue, ColorPeel~\cite{colorpeel} constructs a synthetic dataset and decouples color and shape during training, facilitating the learning of specific color tokens for precise color control. In visual text generation, the use of synthetic data for customizing text attributes is also an intuitive approach. Notably, DiffSTE~\cite{diffste} and Glyph-SDXL~\cite{glyphbyt5} allow for the integration of font and color names directly into prompts, enabling diffusion models to learn these concepts and facilitating precise control over text attributes during inference.
\section{Methodology}
\label{sec:method}

\begin{figure*}[htbp]
 \centering
 \vspace{0mm}
 \includegraphics[width=1.0\textwidth]{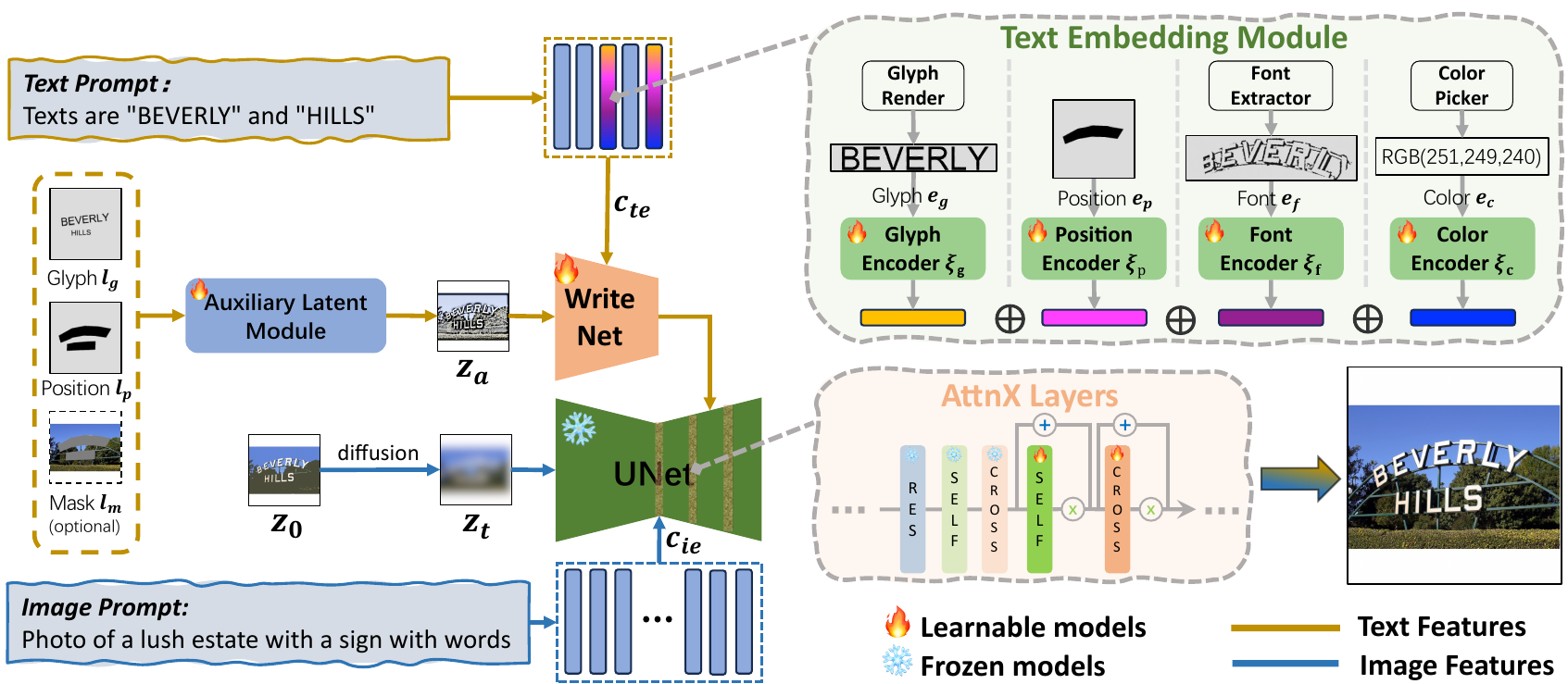}
 \caption{The framework of AnyText2, which is designed with a WriteNet+AttnX architecture to integrate text generation capability into pre-train diffusion models, and there is a Text Embedding Module to provide various conditional control for text generation.}
 \label{fig:framework}
\end{figure*}

AnyText2 is designed as a plugin for pretrained T2I models to enhance their text rendering capabilities. The framework is depicted in Fig.~\ref{fig:framework}. In the standard Latent Diffusion Model (LDM)~\cite{Rombach_LDM_CVPR22}, the original latent pixels $z_{0}$ are gradually adding noise $\epsilon$ through a forward diffusion process to obtain a noisy latent pixels $z_{t}$. The image prompt is then encoded into conditional embeddings $c_{ie}$ using a pre-trained CLIP~\cite{CLIP} text encoder. Both $z_{t}$ and $c_{ie}$ are then fed into a conditional U-Net~\cite{UNET} denoiser $\epsilon_{\theta}$ to predict the noise. The final image is generated after $t$ time steps of the reverse denoising process. To enhance text rendering capabilities, we utilize an Auxiliary Latent Module, similar to AnyText, which encodes the glyph, position, and optionally a masked image (to enable text editing), producing auxiliary pixels $z_a$. The text prompt is processed through a Text Embedding Module to obtain the conditional embeddings $c_{te}$. This Module comprises multiple encoders designed to facilitate various conditional controls. Both $(z_{t}, c_{ie})$ and $(z_{a}, c_{te})$ undergo cross-attention computations in U-Net and WriteNet to better guide the image and text generation, respectively. The integration of image and text is then performed through the U-Net decoder with inserted AttnX layers.  More formally, the optimization objective of our method is represented by the following equation:
\begin{equation}
\mathcal{L}=\mathbb{E}_{\textbf{\textit{z}}_0, \textbf{\textit{c}}_{ie}, \textbf{\textit{z}}_a, \textbf{\textit{c}}_{te}, \textbf{\textit{t}}, \epsilon\sim\mathcal{N}(0,1)}\left[\|\epsilon-\epsilon_{\theta}(\textbf{\textit{z}}_t, \textbf{\textit{z}}_a, \textbf{\textit{c}}_{te}, \textbf{\textit{c}}_{ie}, \textbf{\textit{t}})\|_{2}^{2}\right]  \\
\end{equation}
Next, we will provide a detailed introduction to the key components.

\subsection{WriteNet+AttnX}

\begin{figure*}[htbp]
 \centering
 \vspace{0mm}
 \includegraphics[width=0.8\textwidth]{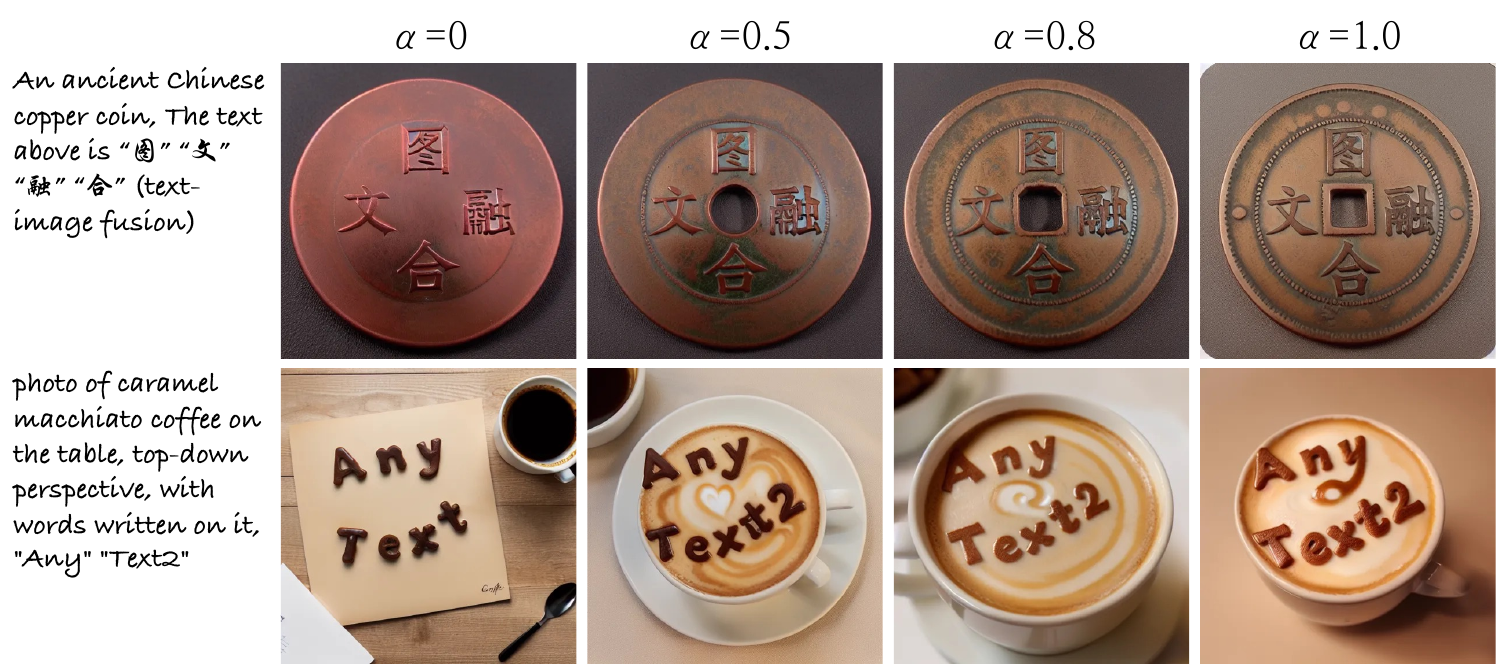}
 \caption{By adjusting the strength coefficient $\alpha$ from 0 to 1 shows that the text-image fusion is gradually improving.}
 \label{fig:attnx_scale}
\end{figure*}

Currently, numerous works~\cite{Yang_GlyphControl_Corr23, zhang2023_brushyourtext, tuo2023anytext, glyphdraw2} utilize a ControlNet-like module as a core component for text generation. Typically, the input to the ControlNet includes not only conditional signals but also noisy latent images and captions describing the image content. In other words, the module is responsible for generating image content in collaboration with the U-Net rather than merely encoding text information (as analyzed in Appendix~\ref{app: analysis}). This method presents two significant issues. First, the training data for text generation tasks often contains a considerable amount of low-quality images with chaotic text, and involving the learnable ControlNet-like module in image generation may potentially degrade image quality. Second, encoding identical text information (glyphs, positions, etc.) at each time step can reduce inference efficiency. To address these concerns, we have transformed the ControlNet-like module into WriteNet by removing the timestep layers and eliminating any image-related inputs, including noisy latent and descriptions of image content in the prompt, thus conducting inference only once. WriteNet focuses exclusively on learning how to write text, while the production-ready U-Net, trained on billions of images, is solely responsible for generating the image content.

To achieve seamless blending of the intermediate text features with the image content, it is essential for them to adequately interact with image latent pixels and conditional embeddings. This interaction can be facilitated through the self-attention and cross-attention mechanisms. Therefore, we insert trainable self-attention and cross-attention layers, denoted as AttnX layers, within the attention blocks in the middle and decoder part of the U-Net. The output from each AttnX layer is multiplied by a strength coefficient $\alpha$ and combined with the output from the previous layer via a shortcut connection. By adjusting $\alpha$, we can modulate the fusion strength between text and image, as illustrated in Fig.~\ref{fig:attnx_scale}. Similarly, the output from WriteNet is also multiplied by a strength coefficient $\beta$ to regulate the intensity of the injected intermediate text features. Notably, setting both $\alpha=0$ and $\beta=0$ allows AnyText2 to generate images without text, relying solely on the original diffusion model.

\subsection{Text Embedding Module}
\label{sec:text_embedding_module}
The Text Embedding Module incorporates four encoders, denoted as $\xi_{g}$, $\xi_{p}$, $\xi_{f}$, and $\xi_{c}$, which are designed to encode the glyph image $e_{g}$, position image $e_{p}$, font image $e_{f}$, and text color $e_{c}$, respectively. The features output by each encoder are mapped to embeddings with the same dimension, which are then summed to produce a representation $r_{i}$ that effectively captures the attributes of the i-th text line:
\begin{equation}
r_i = \xi_{g}(e_{g}) + \xi_{p}(e_{p}) + \xi_{f}(e_{f}) + \xi_{c}(e_{c})
\end{equation}
For the text prompt $y_{t}$, each text line is replaced with a special placeholder $S_*$. After performing tokenization and embedding lookup, denoted as $\phi(\cdot)$, embeddings of all tokens are obtained. We then substitute the attribute representations of $n$ text lines back at $S_*$ and utilize the CLIP text encoder $\tau_{\theta}$ to generate the final conditional embeddings $c_{te}$:
\begin{equation}
\textbf{\textit{c}}_{te}=\tau_{\theta}(\phi({y_{t}}), r_0, r_1, ..., r_{n-1})
\end{equation}
For the glyph encoder $\xi_{g}$, we employ PP-OCRv3~\cite{chen_ppocrv3_corr22} to extract the glyph features as done in AnyText. Next, we will provide a detailed introduction to the other encoders.

\subsubsection{Position Encoder}
In the Cross-Attention mechanisms, the attention map is computed as:
\begin{equation}
\mathcal{M}=Softmax(\frac{QK^T}{\sqrt{d}})
\label{equ:attnmap}
\end{equation}
Intuitively, $\mathcal{M}$ reflects the similarity between Q and K. More specifically, in our context, the element $\mathcal{M}_{ij}$ defines the weight of the $j$-th conditional embedding on the $i$-th auxiliary pixel. However, the embedding of a particular text line is not explicitly associated with the pixels corresponding to its text area. Therefore, we introduce a position encoder $\xi_{p}$ that employs four stacked convolutional layers to encode the position image $e_p$, followed by an average pooling layer. By utilizing $\xi_{p}$, we introduce spatial information for each text line, enabling the embedding to achieve spatial awareness. In the ablation study in Sec.~\ref{sec:acc_real}, we demonstrate that this significantly improves the text accuracy.

\subsubsection{Font Encoder}
Instead of striving for precise separation of text from the complex background, we employ a straightforward adaptive threshold on the text regions to construct a font extractor, resulting in a rough binary image that serves as the font image $e_f$. Examples can be found in Appendix~\ref{app: font_color}. We utilize PP-OCRv3~\cite{chen_ppocrv3_corr22}, just as in the glyph encoder $\xi_{g}$, to construct our font encoder $\xi_{f}$; the only difference is that the OCR model is set to be trainable. The rationale behind this design is that the OCR model inherently focuses on the text portions despite a noisy background. Additionally, the OCR model naturally perceives font types and is trained to be invariant to font variations. To shift its attention from glyphs to fonts, we allow the parameters to be trainable throughout the training process. Our attempts have revealed that this design excels at encoding font styles, while other network architectures, such as convolutional layers or the pre-trained DINOv2~\cite{dinov2}, struggle to encode these noisy font images. During inference, to construct $e_f$, we can either render the text using a user-specified font or select a text region from an image and input it into the font extractor. More examples are illustrated in Fig.~\ref{fig:attribute_customization}. Notably, incorporating font style features into the conditional embeddings enhances the similarity between Q and K in Equation~\ref{equ:attnmap}, which in turn improves text accuracy, as detailed in Sec.~\ref{sec:acc_real}.

\subsubsection{Color Encoder}
\begin{figure*}[htbp]
 \centering
 \vspace{0mm}
 \includegraphics[width=0.8\textwidth]{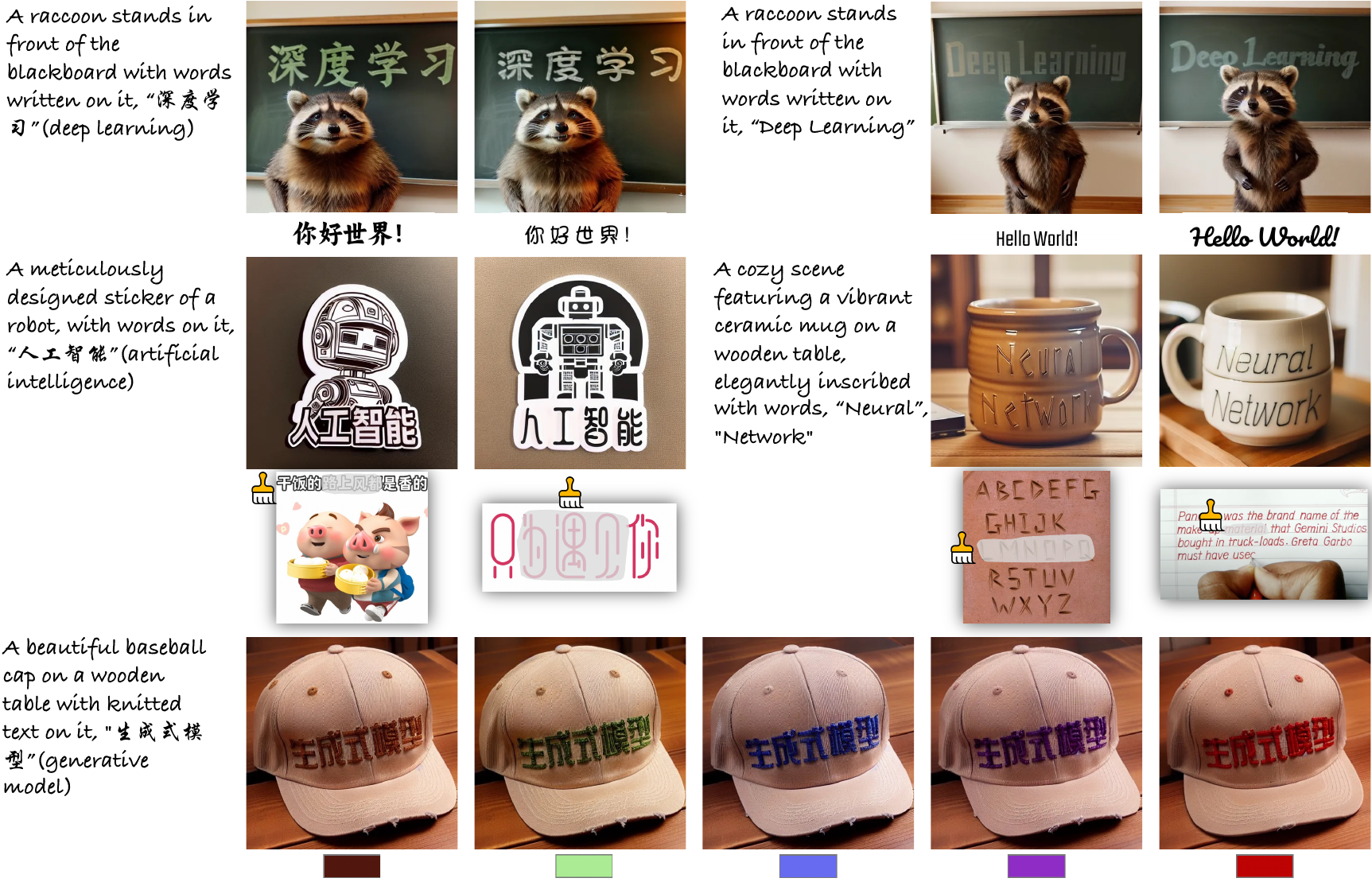}
 \caption{Examples of customizing text attributes. The first row demonstrates font style control using a user-specified font file. The second row showcases selecting a text region from an image to mimic its font style. The third row illustrates the control of text color.}
 \label{fig:attribute_customization}
\end{figure*}

We employ a non-learning method to create a color picker for obtaining the RGB labels of the text. Initially, the colors of all pixels within the text region are clustered and ranked, from which we select the top dominant color blocks. We then analyze their shapes and positions using morphological analysis techniques to identify the most likely text blocks, outputting the mean RGB value as the text color $e_c$. According to our statistics, approximately 65\% text lines in the training data conform to specific criteria, yielding reliable color labels with an accuracy exceeding 90\%. Examples can be found in Appendix~\ref{app: font_color}. In constructing the color encoder $\xi_c$, we experimented with various techniques, including Fourier feature encoding and convolutional layers. However, our attempts have revealed that a simple linear projection layer is sufficient to encode the three RGB values. More examples are illustrated in Fig.~\ref{fig:attribute_customization}.

\section{Dataset and Benchmark}
\label{sec:dataset}
We utilize AnyWord-3M~\cite{tuo2023anytext}, a large-scale multilingual dataset as our training dataset. The dataset contains 3.53 million images, representing a diverse array of scenes containing text, such as street views, book covers, advertisements, posters, and movie frames. However, the captions in AnyWord-3M were generated by BLIP-2~\cite{Li_BLIP2_Corr23}, which lack detailed and accurate descriptions. To improve this, we regenerated the captions using QWen-VL~\cite{Qwen-VL}. Statistics analysis reveals that the BLIP-2 captions contain only 8 words on average, while those generated by QWen-VL are around 47 words. This substantial increase in caption length significantly enhances the description of image details. In the ablation study detailed in Sec.~\ref{sec:long_caption}, we found that while longer captions slightly reduce text accuracy, they significantly improve the model's prompt-following ability. Thus, we opted to train with the longer captions. Examples of training images and captions can be found in Appendix~\ref{app:captions}.

We use the AnyText-benchmark to evaluate the performance of the model, which includes 1,000 images extracted from Wukong~\cite{wukong_corr22} and 1000 images from LAION-400M~\cite{laion_400m_corr21}. This benchmark quantitatively assesses the model's performance in Chinese and English generation, respectively. The benchmark employs three evaluation metrics: Sentence Accuracy (Sen.ACC) and Normalized Edit Distance (NED) for measuring text accuracy using the DuGangOCR~\cite{DuGuangOCR} model, as well as the Frechet Inception Distance (FID) for assessing image authenticity. In addition to these, we incorporate CLIPScore~\cite{clipscore} to evaluate the model's prompt-following capability.

\section{Experiments}
\subsection{Implementation Details}
\label{subsec:details}
We use SD1.5\footnote{https://huggingface.co/runwayml/stable-diffusion-v1-5} to initialize the model weights. The model was trained for 10 epochs on AnyWord-3M using 8 Tesla A100 GPUs, taking approximately two weeks. The AdamW optimizer is employed with a learning rate of 2e-5 and a batch size of 48. The resolutions of $l_g$, $l_p$, $l_m$, and $e_p$ are 512x512, while the resolutions of $e_g$ and $e_f$ are 80x512. The strength coefficients $\alpha$ and $\beta$ are configured to 1.0. A probability of 50\% is applied to choose between inputting $l_m$ or an empty image, facilitating training for both text generation and editing. A probability of 20\% is used to input an empty $e_f$, enabling the model to generate random fonts when no font style is specified. Approximately 35\% of $e_c$ are assigned a default value due to the absence of color labels, allowing for the generation of text in random colors when no color is specified. For image prompts that exceed the maximum length of CLIP, they are split into chunks, processed separately, and then concatenated. For text prompts, we randomly select a phrase from a template and concatenate it with the text content, such as ‘Image with words...' or ‘Text says...'. Compared to AnyText, we have only increased the parameters by 4.5\%(63.8M), while the design of WriteNet+AttnX architecture has improved the inference speed by 19.8\%, as detailed in Appendix~\ref{app: param_size}. 

\subsection{Comparison results}

    \subsubsection{Quantitative Results}
We evaluated ControlNet~\cite{Zhang_ControlNet_Corr23}, TextDiffuser~\cite{Chen_TextDiffuser_Corr23}, GlyphControl~\cite{Yang_GlyphControl_Corr23}, AnyText~\cite{tuo2023anytext}, and GlyphDraw2~\cite{glyphdraw2} using the benchmarks and metrics outlined in Sec.~\ref{sec:dataset}. To ensure a fair evaluation, all publicly available methods employed the DDIM sampler with 20 sampling steps, a CFG scale of 9, a fixed random seed of 100, a batch size of 4, and consistent positive and negative prompt words. The quantitative comparison results are presented in Table~\ref{table:quanti_res}. For GlyphDraw2, we referenced the metrics reported in their paper, achieving a 7.27\% improvement in English Sentence Accuracy (Sen. ACC). A comparison for Chinese was not included as they utilized the PWAcc metric and insufficient details were provided. Notably, AnyText2 outperformed AnyText across all evaluation metrics, particularly in the long caption scenario, where it improved English and Chinese Sen. ACC by 9.3\% and 3.3\%, respectively. Furthermore, it demonstrated significant enhancements in image realism (FID) and prompt-following (CLIPScore).

\begin{table*}
    \vspace{0mm}
    \small
    \centering
    \caption{\small Quantitative comparison of AnyText2 and competing methods. \dag is trained on LAION-Glyph-10M, and \ddag is fine-tuned on TextCaps-5k. Numbers in \brown{brown} color represent the results obtained using the long caption version of the AnyText-benchmark.}
    \vspace{0.0in}
    \setlength{\tabcolsep}{1.5mm}{
    \begin{tabular}{l|c|c|c|c|c|c|c|c}
    \hline
    \multirow{2}{*}{Methods}   & \multicolumn{4}{c|}{English}  & \multicolumn{4}{c}{Chinese}  \\
    \cline{2-9}  & Sen.ACC↑ & NED↑ & FID↓ & CLIPScore↑ & Sen.ACC↑ & NED↑ & FID↓ & CLIPScore↑ \\ \hline
    ControlNet   & 0.5837 & 0.8015 & 45.41 & 0.8448 & 0.3620 & 0.6227 & 41.86 & 0.7801 \\
    TextDiffuser  & 0.5921 & 0.7951 & 41.31 & 0.8685 & 0.0605 & 0.1262 & 53.37 & 0.7675 \\
    GlyphControl\dag & 0.3710 & 0.6680 & 37.84 & 0.8847 & 0.0327 & 0.0845 & 34.36 & 0.8048 \\
    GlyphControl\ddag & 0.5262 & 0.7529 & 43.10 & 0.8548 & 0.0454 & 0.1017 & 49.51 & 0.7863 \\
    \multirow{2}{*}{Anytext} & 0.7239 & 0.8760 & 33.54 & 0.8841 & 0.6923 & 0.8396 & 31.58 & 0.8015 \\
                & \brown{0.7242} & \brown{0.8780} & \brown{35.27} & \brown{0.9602} & \brown{0.6917} & \brown{0.8373} & \brown{31.38} & \brown{0.8870} \\
    GlyphDraw2 & 0.7369 & 0.8921 & - & - & - & - & - & - \\
    \hline
    \multirow{2}{*}{Anytext2} & \textbf{0.8096} & \textbf{0.9184} & \textbf{33.32} & \textbf{0.8963} & \textbf{0.7130} & \textbf{0.8516} & \textbf{27.94} & \textbf{0.8139} \\
                 & \brown{\textbf{0.8175}} & \brown{\textbf{0.9193}} & \brown{\textbf{27.87}} & \brown{\textbf{0.9882}} & \brown{\textbf{0.7250}} & \brown{\textbf{0.8529}} & \brown{\textbf{24.32}} & \brown{\textbf{0.9137}} \\
    \hline
\end{tabular}}
    \vspace{-0.15in}
    \label{table:quanti_res}
\end{table*}

    \subsubsection{Qualitative Results}

\begin{figure*}[htbp]
 \centering
 \vspace{0mm}
 \includegraphics[width=0.9\textwidth]{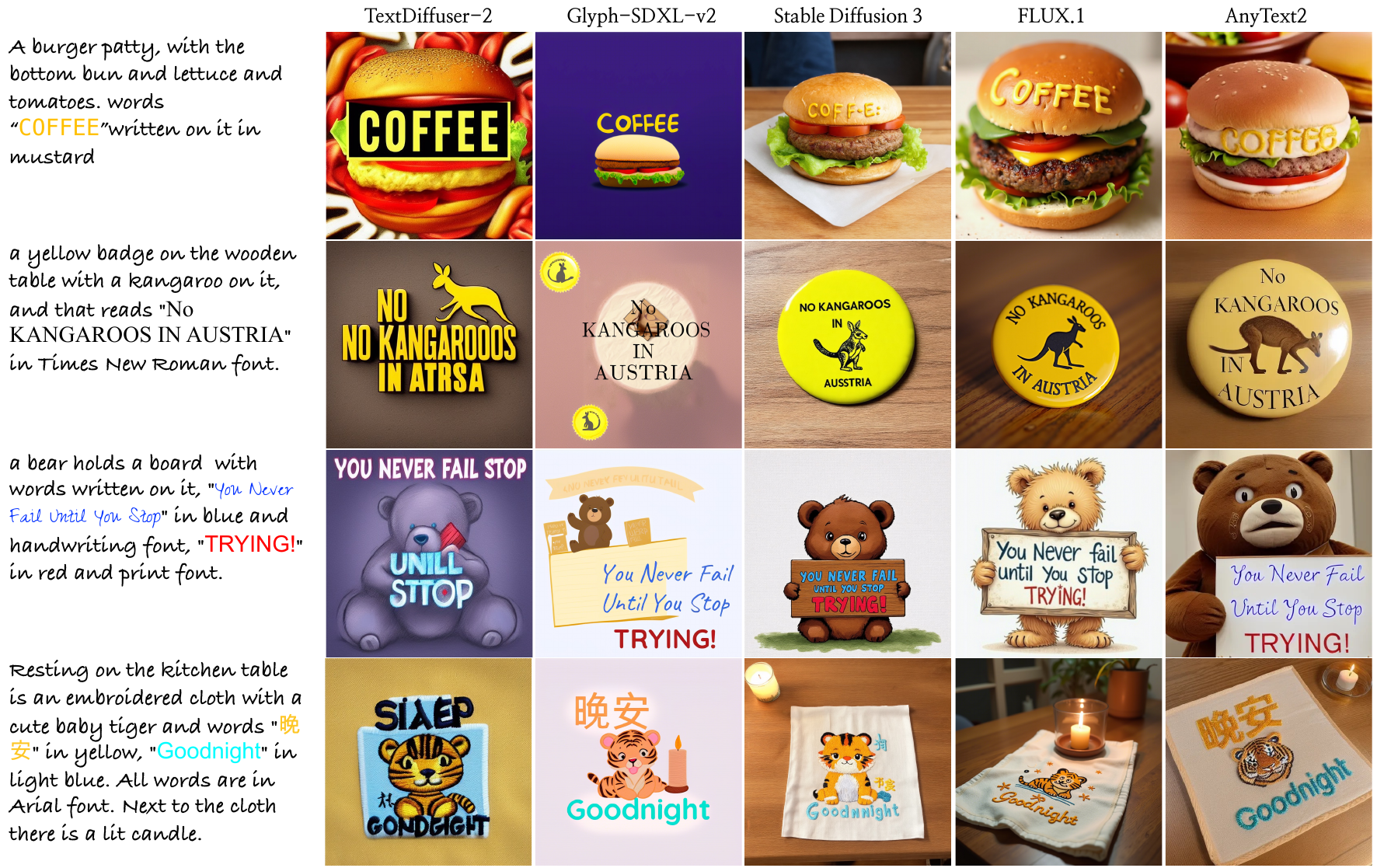}
 \caption{Qualitative comparison of AnyText2 and other methods. From the perspectives of text accuracy, text-image integration, attribute customization, and multilingual support, AnyText2 demonstrated significant advantages.}
 \label{fig:qualitative}
\end{figure*}

As shown in Fig.~\ref{fig:qualitative}, we conducted a qualitative comparison of AnyText2 with several recent methods, including TextDiffuser-2~\cite{Chen_TextDiffuser2}, Glyph-SDXL-v2~\cite{glyphbyt5-v2}, Stable Diffusion 3~\cite{Patrick_SD3_ICML24}, and FLUX.1~\cite{FLUX.1}. The image captions in the leftmost column are input directly to TextDiffuser-2, SD3, and FLUX.1. For Glyph-SDXL-v2 and AnyText2, the inputs were adjusted according to each method's requirements, which included manually setting the layout and selecting appropriate text fonts or colors based on the captions. Each method underwent multiple trials, and we present one of the best results. From the results, TextDiffuser-2 demonstrates subpar performance in text accuracy, particularly when handling multiple lines. Glyph-SDXL-v2 achieves good accuracy and allows for precise customization of text fonts and colors in multilingual text; however, it is limited to generating overlaid text on images, showing little correlation to the image content. Although SD3 produces visually appealing images, its accuracy in English is moderate, offering only rough control over colors and minimal control over fonts. FLUX.1 generates impressive visual results while maintaining decent English accuracy, though it suffers from occasional capitalization errors. Unfortunately, it only permits rough control over simple fonts and colors in the prompt and is limited to the English generation. In comparison, AnyText2 stands out with the best accuracy, the most precise control over text attributes, seamless text-image integration, and robust multilingual support.

\subsection{Ablation study}
\label{sec:ablation_study}
We extracted 200k images from AnyWord-3M, which includes 100k images each for Chinese and English, and conducted ablation experiments by training on this small-scale dataset for 15 epochs to validate each module of our method. Next, we will analyze our method from two perspectives: accuracy and realism, as well as prompt-following capability.

\subsubsection{Accuracy and Realism}
\label{sec:acc_real}

\begin{table*}
        \small
    \centering
    \caption{\small Ablation experiments of AnyText2 on a small-scale dataset from AnyWord-3M.}
    \vspace{0.0in}
    \setlength{\tabcolsep}{1.35mm}{
    \begin{tabular}{c|c|c|c|c|c|c|c|c|c|c|c|c|c}
    \hline
    \multirow{2}{*}{Exp.} & \multirow{2}{*}{Pos.} & \multirow{2}{*}{Font} & \multirow{2}{*}{Color} & \multicolumn{3}{c|}{\multirow{1}{*}{AttnX}} & \multirow{2}{*}{WriteNet} & \multicolumn{3}{c|}{English} & \multicolumn{3}{c}{Chinese} \\  
    \cline{5-7} \cline{9-14} & & & & 2 & 1 & 0 & & Sen.ACC↑ & NED↑ &FID↓ &Sen.ACC↑ & NED↑ &FID↓  \\ \hline
    1    &  &  &   &   &   &   &   &                0.4873 & 0.7721 & 35.38 & 0.5404 & 0.7631 & 31.19   \\
    2    & \checkmark &  &  &  &  &  &  &           0.5237 & 0.7876 & 35.86 & 0.5681 & 0.7725 & 29.45 \\
    3    & \checkmark & \checkmark &   & & & & &    0.5926 & 0.8276 & 38.57 & 0.5688 & 0.7763 & 31.60 \\
    4    & \checkmark & \checkmark & \checkmark & & & & &         0.5732 & 0.8169 & 37.24 & 0.5525 & 0.7620 & 32.69  \\
    5    & \checkmark & \checkmark & \checkmark & \checkmark & \checkmark & \checkmark &  &     0.6372 & 0.8481 & 44.98 & 0.5632 & 0.7769 & 35.84  \\
    6    & \checkmark & \checkmark & \checkmark & \checkmark & \checkmark &  &   &  \textbf{0.6391} & \textbf{0.8490} & 39.14 & \textbf{0.5760} & \textbf{0.7858} & 32.92  \\
    7    & \checkmark & \checkmark & \checkmark & \checkmark &  &  &   &  0.6343 & 0.8478 & 37.21 & 0.5527 & 0.7696 & \textbf{28.91}  \\
    8    & \checkmark & \checkmark & \checkmark & \checkmark & \checkmark &  & \checkmark &    0.6335 & 0.8443 & \textbf{35.19} & 0.5614 & 0.7731 & 29.40  \\ \hline
       \end{tabular}}
    \vspace{-3mm}
        \label{table:ablation}
    
\end{table*}

In Table~\ref{table:ablation}, we validated the effectiveness of each module in AnyText2. In Exp.1, the original AnyText serves as the baseline. Incrementally adding the position and font encoders in the Text Embedding Module during Exp.2\&3 significantly boosted text accuracy, attributed to improved similarity between auxiliary pixels and conditional embeddings. In Exp.4, adding the color encoder resulted in a slight decline in accuracy, possibly due to incorrect ground truths in color labels and challenges in learning text colors against complex backgrounds. Exp.5-7 demonstrated that AttnX layers improved text accuracy, with their position affecting the FID score. Specifically, inserting AttnX closer to the U-Net output led to a tendency for it to directly copy glyphs from the auxiliary pixels due to U-Net's skip connections rather than learning to generate them. Thus, we inserted AttnX in the first two blocks of the U-Net decoder for optimal accuracy and realism. In Exp. 8, replacing the ControlNet-like module with WriteNet slightly decreased accuracy but significantly improved the FID score. This aligns with our expectation of a trade-off between image realism and text accuracy, as embedded text is more challenging for OCR than overlaid text despite being more realistic. Considering accuracy, realism, and inference efficiency, we selected the configuration from Exp.8 for training on the full dataset.

\subsubsection{Prompt-Following}
\label{sec:long_caption}

\begin{table}
        \small
    \centering
    \caption{\small Ablation experiments of training AnyText2 using both short(6S) and long captions(6L), evaluated with the the AnyText-benchmark under both short and long caption (marked in \brown{brown}) scenarios. Eps denotes Epochs, S.A denotes Sen.ACC, and C.P denotes CLIPScore.}
    \vspace{0.0in}
    \setlength{\tabcolsep}{1.35mm}{
    \begin{tabular}{c|c|c|c|c|c|c|c}
    \hline
    \multirow{2}{*}{Exp.} & \multirow{2}{*}{Eps}& \multicolumn{3}{c|}{English} & \multicolumn{3}{c}{Chinese} \\  
    \cline{3-8} & & S.A↑ & NED↑ &C.S↑ &S.A↑ & NED↑ &C.S↑  \\ \hline
    6S  &   \multirow{2}{*}{15}  &           0.6391 & 0.8490 & 0.8797 & 0.5760 & 0.7858 & 0.7941   \\
    6L  &    &       0.6094 & 0.8296 & 0.8734 & 0.4995 & 0.7401 & 0.7952 \\
    \cline{1-8}  6S  &    \multirow{2}{*}{19} & 0.6313 & 0.8459 & 0.8828 & 0.5719 & 0.7830 & 0.7948 \\
    6L  &    &     0.6182 & 0.8360 & 0.8773 & 0.5541 & 0.7710 & 0.8036  \\
    \cline{1-8}  6S  &     \multirow{2}{*}{15} & \brown{0.6479} & \brown{0.8481} & \brown{0.8577} & \brown{0.5606} & \brown{0.7738} & \brown{0.7333}  \\
    6L  &  &  \brown{0.6305} & \brown{0.8412} & \brown{0.8650} & \brown{0.5055} & \brown{0.7422} & \brown{0.7511}  \\
    \cline{1-8}  6S  &  \multirow{2}{*}{19} &  \brown{0.6453} & \brown{0.8476} & \brown{0.8596} & \brown{0.5639} & \brown{0.7784} & \brown{0.7372}  \\
    6L  &  &  \brown{0.6357} & \brown{0.8431} & \brown{0.8665} & \brown{0.5618} & \brown{0.7738} & \brown{0.7542}  \\ \hline
       \end{tabular}}
    \vspace{-3mm}
        \label{table:long_caption}
    
\end{table}

We trained two models in the configuration of Exp.6 using short (6S) and long (6L) captions to examine the impact of caption length on accuracy and prompt-following, as shown in Table~\ref{table:long_caption}. The first two rows indicate a significant accuracy drop with long captions, particularly a 7.6\% decrease in Chinese Sen. ACC. This decline was partly due to the model with long captions not fully converging on the small-scale training set. After continuing training for an additional 4 epochs, Exp.6S metrics reached saturation while those for Exp.6L improved. Although the performance gap narrowed, Exp.6L's CLIPScore remained comparable. We then replaced the AnyText benchmark with long captions, observing a similar trend. After 19 epochs, the accuracy gap further diminished, but Exp.6L's CLIPScore was significantly higher than Exp.6S. These findings suggest that training with long captions may slightly decrease text accuracy while enhancing prompt-following capabilities, particularly for complex captions. Therefore, we opted to use long captions for training on the full dataset.

\section{Conclusion}
In this paper, we introduced AnyText2, a novel method that tackles the cutting-edge challenge of precisely controlling text attributes in realistic image generation. We explored techniques for extracting font and color labels from natural scene images and developed dedicated encoders for feature representation, enabling the customization of text attributes for each line. Additionally, we conducted an in-depth analysis of visual text generation mechanisms and creatively proposed the WriteNet+AttnX architecture, which decouples text and image generation tasks while effectively integrating them through attention layers. Our approach outperformed its predecessor, AnyText, achieving higher accuracy, enhanced realism, and faster inference speed. Furthermore, the model's prompt-following capabilities were bolstered through the use of long captions. In future work, we will continue to push the boundaries of visual text generation and aim to gradually port AnyText2 to more innovative T2I models.
{
    \small
    \bibliographystyle{ieeenat_fullname}
    \bibliography{main}

\begin{thebibliography}{47}
\providecommand{\natexlab}[1]{#1}
\providecommand{\url}[1]{\texttt{#1}}
\expandafter\ifx\csname urlstyle\endcsname\relax
  \providecommand{\doi}[1]{doi: #1}\else
  \providecommand{\doi}{doi: \begingroup \urlstyle{rm}\Url}\fi

\bibitem[Bai et~al.(2023)Bai, Bai, Yang, Wang, Tan, Wang, Lin, Zhou, and Zhou]{Qwen-VL}
Jinze Bai, Shuai Bai, Shusheng Yang, Shijie Wang, Sinan Tan, Peng Wang, Junyang Lin, Chang Zhou, and Jingren Zhou.
\newblock Qwen-vl: A versatile vision-language model for understanding, localization, text reading, and beyond.
\newblock \emph{arXiv preprint arXiv:2308.12966}, 2023.

\bibitem[BlackForestLab(2024)]{FLUX.1}
BlackForestLab.
\newblock Flux.1.
\newblock \url{https://blackforestlabs.ai/announcing-black-forest-labs/}, 2024.

\bibitem[Butt et~al.(2024)Butt, Wang, Vazquez-Corral, and van~de Weijer]{colorpeel}
Muhammad~Atif Butt, Kai Wang, Javier Vazquez-Corral, and Joost van~de Weijer.
\newblock Colorpeel: Color prompt learning with diffusion models via color and shape disentanglement, 2024.

\bibitem[Chen et~al.(2023{\natexlab{a}})Chen, Huang, Lv, Cui, Chen, and Wei]{Chen_TextDiffuser2}
Jingye Chen, Yupan Huang, Tengchao Lv, Lei Cui, Qifeng Chen, and Furu Wei.
\newblock Textdiffuser-2: Unleashing the power of language models for text rendering.
\newblock \emph{arXiv preprint arXiv:2311.16465}, 2023{\natexlab{a}}.

\bibitem[Chen et~al.(2023{\natexlab{b}})Chen, Huang, Lv, Cui, Chen, and Wei]{Chen_TextDiffuser_Corr23}
Jingye Chen, Yupan Huang, Tengchao Lv, Lei Cui, Qifeng Chen, and Furu Wei.
\newblock Textdiffuser: Diffusion models as text painters.
\newblock \emph{arXiv preprint}, abs/2305.10855, 2023{\natexlab{b}}.

\bibitem[Esser et~al.(2024)Esser, Kulal, Blattmann, Entezari, M{\"{u}}ller, Saini, Levi, Lorenz, Sauer, Boesel, Podell, Dockhorn, English, and Rombach]{Patrick_SD3_ICML24}
Patrick Esser, Sumith Kulal, Andreas Blattmann, Rahim Entezari, Jonas M{\"{u}}ller, Harry Saini, Yam Levi, Dominik Lorenz, Axel Sauer, Frederic Boesel, Dustin Podell, Tim Dockhorn, Zion English, and Robin Rombach.
\newblock Scaling rectified flow transformers for high-resolution image synthesis.
\newblock In \emph{ICML}, 2024.

\bibitem[Gal et~al.(2023)Gal, Alaluf, Atzmon, Patashnik, Bermano, Chechik, and Cohen{-}Or]{Gal_Inversion_ICLR23}
Rinon Gal, Yuval Alaluf, Yuval Atzmon, Or Patashnik, Amit~Haim Bermano, Gal Chechik, and Daniel Cohen{-}Or.
\newblock An image is worth one word: Personalizing text-to-image generation using textual inversion.
\newblock In \emph{ICLR}, 2023.

\bibitem[Goodfellow et~al.(2014)Goodfellow, Pouget{-}Abadie, Mirza, Xu, Warde{-}Farley, Ozair, Courville, and Bengio]{Goodfellow_GAN_NPIS14}
Ian~J. Goodfellow, Jean Pouget{-}Abadie, Mehdi Mirza, Bing Xu, David Warde{-}Farley, Sherjil Ozair, Aaron~C. Courville, and Yoshua Bengio.
\newblock Generative adversarial nets.
\newblock In \emph{NeurIPS}, 2014.

\bibitem[Gu et~al.(2022)Gu, Meng, Lu, Hou, Niu, Xu, Liang, Zhang, Jiang, and Xu]{wukong_corr22}
Jiaxi Gu, Xiaojun Meng, Guansong Lu, Lu Hou, Minzhe Niu, Hang Xu, Xiaodan Liang, Wei Zhang, Xin Jiang, and Chunjing Xu.
\newblock Wukong: 100 million large-scale chinese cross-modal pre-training dataset and {A} foundation framework.
\newblock \emph{CoRR}, abs/2202.06767, 2022.

\bibitem[He et~al.(2024)He, Chen, Wang, Liu, Du, Tao, and Qiao]{He_DiffFont_IJCV24}
Haibin He, Xinyuan Chen, Chaoyue Wang, Juhua Liu, Bo Du, Dacheng Tao, and Yu Qiao.
\newblock Diff-font: Diffusion model for robust one-shot font generation.
\newblock \emph{IJCV}, abs/2212.05895, 2024.

\bibitem[Hertz et~al.(2023)Hertz, Mokady, Tenenbaum, Aberman, Pritch, and Cohen{-}Or]{Hertz_prompt2promt_ICLR23}
Amir Hertz, Ron Mokady, Jay Tenenbaum, Kfir Aberman, Yael Pritch, and Daniel Cohen{-}Or.
\newblock Prompt-to-prompt image editing with cross-attention control.
\newblock In \emph{ICLR}, 2023.

\bibitem[Hessel et~al.(2021)Hessel, Holtzman, Forbes, Bras, and Choi]{clipscore}
Jack Hessel, Ari Holtzman, Maxwell Forbes, Ronan~Le Bras, and Yejin Choi.
\newblock Clipscore: {A} reference-free evaluation metric for image captioning.
\newblock In \emph{EMNLP}, 2021.

\bibitem[Ho et~al.(2020)Ho, Jain, and Abbeel]{Ho_DDPM_NIPS20}
Jonathan Ho, Ajay Jain, and Pieter Abbeel.
\newblock Denoising diffusion probabilistic models.
\newblock In \emph{NeurIPS}, 2020.

\bibitem[Hu et~al.(2023)Hu, Gao, Zhang, Sun, Zhang, and Bo]{hu_animateanyone}
Li Hu, Xin Gao, Peng Zhang, Ke Sun, Bang Zhang, and Liefeng Bo.
\newblock Animate anyone: Consistent and controllable image-to-video synthesis for character animation.
\newblock \emph{CoRR}, 2023.

\bibitem[Huang et~al.(2023)Huang, Chen, Liu, Shen, Zhao, and Zhou]{Huang_Composer_Corr23}
Lianghua Huang, Di Chen, Yu Liu, Yujun Shen, Deli Zhao, and Jingren Zhou.
\newblock Composer: Creative and controllable image synthesis with composable conditions.
\newblock \emph{arXiv preprint}, abs/2302.09778, 2023.

\bibitem[Ji et~al.(2023)Ji, Zhang, Wang, Hou, Zhang, Price, and Chang]{diffste}
Jiabao Ji, Guanhua Zhang, Zhaowen Wang, Bairu Hou, Zhifei Zhang, Brian Price, and Shiyu Chang.
\newblock Improving diffusion models for scene text editing with dual encoders, 2023.

\bibitem[Kingma and Welling(2014)]{Diederik_VAE_ICLR14}
Diederik~P. Kingma and Max Welling.
\newblock Auto-encoding variational bayes.
\newblock In \emph{ICLR}, 2014.

\bibitem[Li et~al.(2022)Li, Liu, Guo, Yin, Jiang, Du, Du, Zhu, Lai, Hu, Yu, and Ma]{chen_ppocrv3_corr22}
Chenxia Li, Weiwei Liu, Ruoyu Guo, Xiaoting Yin, Kaitao Jiang, Yongkun Du, Yuning Du, Lingfeng Zhu, Baohua Lai, Xiaoguang Hu, Dianhai Yu, and Yanjun Ma.
\newblock Pp-ocrv3: More attempts for the improvement of ultra lightweight {OCR} system.
\newblock \emph{CoRR}, abs/2206.03001, 2022.

\bibitem[Li et~al.(2023)Li, Li, Savarese, and Hoi]{Li_BLIP2_Corr23}
Junnan Li, Dongxu Li, Silvio Savarese, and Steven C.~H. Hoi.
\newblock {BLIP-2:} bootstrapping language-image pre-training with frozen image encoders and large language models.
\newblock \emph{arXiv preprint}, abs/2301.12597, 2023.

\bibitem[Li et~al.(2024)Li, Cao, Wang, Qi, Cheng, and Shan]{li2023photomaker}
Zhen Li, Mingdeng Cao, Xintao Wang, Zhongang Qi, Ming-Ming Cheng, and Ying Shan.
\newblock Photomaker: Customizing realistic human photos via stacked id embedding.
\newblock In \emph{IEEE Conference on Computer Vision and Pattern Recognition (CVPR)}, 2024.

\bibitem[Liu et~al.(2023)Liu, Garrette, Saharia, Chan, Roberts, Narang, Blok, Mical, Norouzi, and Constant]{Liu_CharacterAware_ACL23}
Rosanne Liu, Dan Garrette, Chitwan Saharia, William Chan, Adam Roberts, Sharan Narang, Irina Blok, RJ Mical, Mohammad Norouzi, and Noah Constant.
\newblock Character-aware models improve visual text rendering.
\newblock In \emph{ACL}, pages 16270--16297, 2023.

\bibitem[Liu et~al.(2024{\natexlab{a}})Liu, Liang, Liang, Luo, Li, Huang, and Yuan]{glyphbyt5}
Zeyu Liu, Weicong Liang, Zhanhao Liang, Chong Luo, Ji Li, Gao Huang, and Yuhui Yuan.
\newblock Glyph-byt5: A customized text encoder for accurate visual text rendering.
\newblock \emph{arXiv preprint arXiv:2403.09622}, 2024{\natexlab{a}}.

\bibitem[Liu et~al.(2024{\natexlab{b}})Liu, Liang, Zhao, Chen, Li, and Yuan]{glyphbyt5-v2}
Zeyu Liu, Weicong Liang, Yiming Zhao, Bohan Chen, Ji Li, and Yuhui Yuan.
\newblock Glyph-byt5-v2: A strong aesthetic baseline for accurate multilingual visual text rendering.
\newblock \emph{arXiv preprint arXiv:2406.10208}, 2024{\natexlab{b}}.

\bibitem[Ma et~al.(2023)Ma, Zhao, Chen, Wang, Niu, Lu, and Lin]{Ma_glyphdraw_Corr23}
Jian Ma, Mingjun Zhao, Chen Chen, Ruichen Wang, Di Niu, Haonan Lu, and Xiaodong Lin.
\newblock Glyphdraw: Learning to draw chinese characters in image synthesis models coherently.
\newblock \emph{arXiv preprint}, abs/2303.17870, 2023.

\bibitem[Ma et~al.(2024)Ma, Deng, Chen, Lu, and Yang]{glyphdraw2}
Jian Ma, Yonglin Deng, Chen Chen, Haonan Lu, and Zhenyu Yang.
\newblock Glyphdraw2: Automatic generation of complex glyph posters with diffusion models and large language models.
\newblock \emph{CoRR}, 2024.

\bibitem[ModelScope(2023)]{DuGuangOCR}
ModelScope.
\newblock Duguangocr.
\newblock \url{https://modelscope.cn/models/damo/cv_convnextTiny_ocr-recognition-general_damo/summary}, 2023.

\bibitem[Mou et~al.(2023)Mou, Wang, Xie, Zhang, Qi, Shan, and Qie]{Mou_T2I_Corr23}
Chong Mou, Xintao Wang, Liangbin Xie, Jian Zhang, Zhongang Qi, Ying Shan, and Xiaohu Qie.
\newblock T2i-adapter: Learning adapters to dig out more controllable ability for text-to-image diffusion models.
\newblock \emph{arXiv preprint}, abs/2302.08453, 2023.

\bibitem[OpenAI(2023)]{DALLE3}
OpenAI.
\newblock Dall·e3.
\newblock \url{https://openai.com/index/dall-e-3/}, 2023.

\bibitem[Oquab et~al.(2023)Oquab, Darcet, Moutakanni, Vo, Szafraniec, Khalidov, Fernandez, Haziza, Massa, El-Nouby, Howes, Huang, Xu, Sharma, Li, Galuba, Rabbat, Assran, Ballas, Synnaeve, Misra, Jegou, Mairal, Labatut, Joulin, and Bojanowski]{dinov2}
Maxime Oquab, Timothée Darcet, Theo Moutakanni, Huy~V. Vo, Marc Szafraniec, Vasil Khalidov, Pierre Fernandez, Daniel Haziza, Francisco Massa, Alaaeldin El-Nouby, Russell Howes, Po-Yao Huang, Hu Xu, Vasu Sharma, Shang-Wen Li, Wojciech Galuba, Mike Rabbat, Mido Assran, Nicolas Ballas, Gabriel Synnaeve, Ishan Misra, Herve Jegou, Julien Mairal, Patrick Labatut, Armand Joulin, and Piotr Bojanowski.
\newblock Dinov2: Learning robust visual features without supervision, 2023.

\bibitem[Park et~al.(2021{\natexlab{a}})Park, Chun, Cha, Lee, and Shim]{Park_LFFont_AAA21}
Song Park, Sanghyuk Chun, Junbum Cha, Bado Lee, and Hyunjung Shim.
\newblock Few-shot font generation with localized style representations and factorization.
\newblock In \emph{AAAI}, pages 2393--2402, 2021{\natexlab{a}}.

\bibitem[Park et~al.(2021{\natexlab{b}})Park, Chun, Cha, Lee, and Shim]{Park_MXFont_ICCV21}
Song Park, Sanghyuk Chun, Junbum Cha, Bado Lee, and Hyunjung Shim.
\newblock Multiple heads are better than one: Few-shot font generation with multiple localized experts.
\newblock In \emph{ICCV}, pages 13880--13889, 2021{\natexlab{b}}.

\bibitem[Podell et~al.(2024)Podell, English, Lacey, Blattmann, Dockhorn, M{\"{u}}ller, Penna, and Rombach]{Dustin_SDXL_ICLR24}
Dustin Podell, Zion English, Kyle Lacey, Andreas Blattmann, Tim Dockhorn, Jonas M{\"{u}}ller, Joe Penna, and Robin Rombach.
\newblock {SDXL:} improving latent diffusion models for high-resolution image synthesis.
\newblock In \emph{ICLR}, 2024.

\bibitem[Radford et~al.(2021)Radford, Kim, Hallacy, Ramesh, Goh, Agarwal, Sastry, Askell, Mishkin, Clark, Krueger, and Sutskever]{CLIP}
Alec Radford, Jong~Wook Kim, Chris Hallacy, Aditya Ramesh, Gabriel Goh, Sandhini Agarwal, Girish Sastry, Amanda Askell, Pamela Mishkin, Jack Clark, Gretchen Krueger, and Ilya Sutskever.
\newblock Learning transferable visual models from natural language supervision.
\newblock In \emph{ICML}, 2021.

\bibitem[Ramesh et~al.(2021)Ramesh, Pavlov, Goh, Gray, Voss, Radford, Chen, and Sutskever]{Ramesh_DALLE_icml21}
Aditya Ramesh, Mikhail Pavlov, Gabriel Goh, Scott Gray, Chelsea Voss, Alec Radford, Mark Chen, and Ilya Sutskever.
\newblock Zero-shot text-to-image generation.
\newblock In \emph{ICML}, pages 8821--8831. {PMLR}, 2021.

\bibitem[Ramesh et~al.(2022)Ramesh, Dhariwal, Nichol, Chu, and Chen]{Ramesh_DALLE2_corr22}
Aditya Ramesh, Prafulla Dhariwal, Alex Nichol, Casey Chu, and Mark Chen.
\newblock Hierarchical text-conditional image generation with {CLIP} latents.
\newblock \emph{arXiv preprint}, abs/2204.06125, 2022.

\bibitem[Rombach et~al.(2022)Rombach, Blattmann, Lorenz, Esser, and Ommer]{Rombach_LDM_CVPR22}
Robin Rombach, Andreas Blattmann, Dominik Lorenz, Patrick Esser, and Bj\"orn Ommer.
\newblock High-resolution image synthesis with latent diffusion models.
\newblock In \emph{CVPR}, pages 10684--10695, 2022.

\bibitem[Ronneberger et~al.(2015)Ronneberger, Fischer, and Brox]{UNET}
Olaf Ronneberger, Philipp Fischer, and Thomas Brox.
\newblock U-net: Convolutional networks for biomedical image segmentation.
\newblock In \emph{MICCAI}, 2015.

\bibitem[Ruiz et~al.(2022)Ruiz, Li, Jampani, Pritch, Rubinstein, and Aberman]{Ruiz_DreamBooth_Corr22}
Nataniel Ruiz, Yuanzhen Li, Varun Jampani, Yael Pritch, Michael Rubinstein, and Kfir Aberman.
\newblock Dreambooth: Fine tuning text-to-image diffusion models for subject-driven generation.
\newblock \emph{arXiv preprint}, abs/2208.12242, 2022.

\bibitem[Schuhmann et~al.(2021)Schuhmann, Vencu, Beaumont, Kaczmarczyk, Mullis, Katta, Coombes, Jitsev, and Komatsuzaki]{laion_400m_corr21}
Christoph Schuhmann, Richard Vencu, Romain Beaumont, Robert Kaczmarczyk, Clayton Mullis, Aarush Katta, Theo Coombes, Jenia Jitsev, and Aran Komatsuzaki.
\newblock {LAION-400M:} open dataset of clip-filtered 400 million image-text pairs.
\newblock \emph{CoRR}, abs/2111.02114, 2021.

\bibitem[Tuo et~al.(2023)Tuo, Xiang, He, Geng, and Xie]{tuo2023anytext}
Yuxiang Tuo, Wangmeng Xiang, Jun-Yan He, Yifeng Geng, and Xuansong Xie.
\newblock Anytext: Multilingual visual text generation and editing.
\newblock \emph{arXiv}, 2023.

\bibitem[Wang et~al.(2024)Wang, Bai, Wang, Qin, and Chen]{wang2024_instantid}
Qixun Wang, Xu Bai, Haofan Wang, Zekui Qin, and Anthony Chen.
\newblock Instantid: Zero-shot identity-preserving generation in seconds.
\newblock \emph{arXiv preprint arXiv:2401.07519}, 2024.

\bibitem[Yang et~al.(2023)Yang, Gui, Yuan, Ding, Hu, and Chen]{Yang_GlyphControl_Corr23}
Yukang Yang, Dongnan Gui, Yuhui Yuan, Haisong Ding, Han Hu, and Kai Chen.
\newblock Glyphcontrol: Glyph conditional control for visual text generation.
\newblock \emph{arXiv preprint}, abs/2305.18259, 2023.

\bibitem[Yang et~al.(2024)Yang, Peng, Kong, Zhang, Yao, and Jin]{yang2024fontdiffuser}
Zhenhua Yang, Dezhi Peng, Yuxin Kong, Yuyi Zhang, Cong Yao, and Lianwen Jin.
\newblock Fontdiffuser: One-shot font generation via denoising diffusion with multi-scale content aggregation and style contrastive learning.
\newblock In \emph{Proceedings of the AAAI conference on artificial intelligence}, 2024.

\bibitem[Ye et~al.(2023)Ye, Zhang, Liu, Han, and Yang]{ip-adapter}
Hu Ye, Jun Zhang, Sibo Liu, Xiao Han, and Wei Yang.
\newblock Ip-adapter: Text compatible image prompt adapter for text-to-image diffusion models.
\newblock \emph{arXiv preprint}, 2023.

\bibitem[Zhang and Agrawala(2023)]{Zhang_ControlNet_Corr23}
Lvmin Zhang and Maneesh Agrawala.
\newblock Adding conditional control to text-to-image diffusion models.
\newblock \emph{arXiv preprint}, abs/2302.05543, 2023.

\bibitem[Zhang et~al.(2023)Zhang, Chen, Wang, Lu, and Qiao]{zhang2023_brushyourtext}
Lingjun Zhang, Xinyuan Chen, Yaohui Wang, Yue Lu, and Yu Qiao.
\newblock Brush your text: Synthesize any scene text on images via diffusion model, 2023.

\bibitem[Zhao and Lian(2023)]{zhao2023udifftext}
Yiming Zhao and Zhouhui Lian.
\newblock Udifftext: A unified framework for high-quality text synthesis in arbitrary images via character-aware diffusion models, 2023.

\end{thebibliography}
}
\clearpage
\setcounter{page}{1}
\appendix
\renewcommand{\thesection}{\Alph{section}}
\setcounter{section}{0}
\makeatletter
\makeatother

\maketitlesupplementary

\section{Analysis of the Text Generation Process in AnyText}
\label{app: analysis}
\begin{figure*}[htbp]
 \vspace{0mm}
 \centering
 \includegraphics[width=0.8\textwidth]{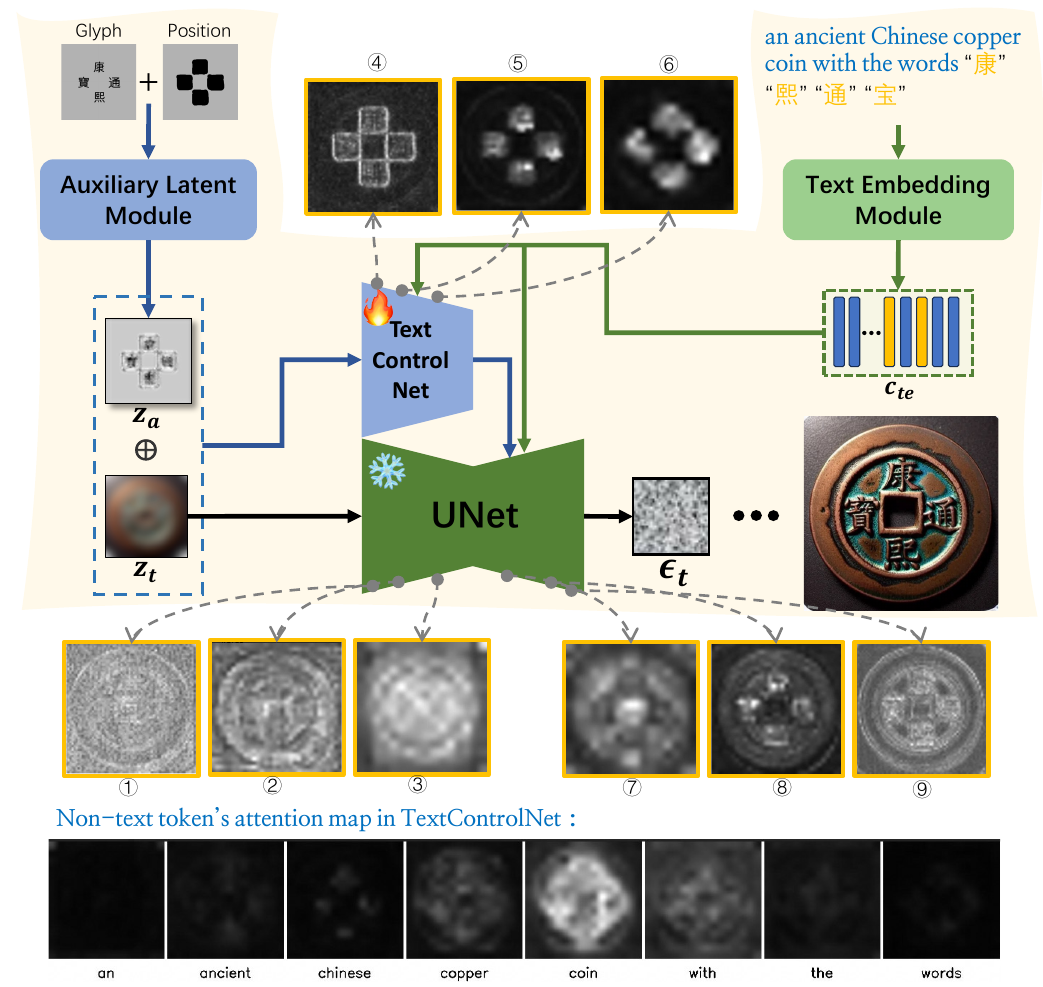}
 \vspace{0mm}
 \caption{Analysis of the text generation process in AnyText.}
 \vspace{0mm}
 \label{fig:analysis}
\end{figure*}

In this section, we examine the text generation process in AnyText~\cite{tuo2023anytext} and visualize the attention maps of its various cross-attention layers using the method from Prompt-To-Prompt~\cite{Hertz_prompt2promt_ICLR23}, as shown in Fig.~\ref{fig:analysis}. AnyText introduces a Text Embedding Module that extracts text glyphs using an OCR model and then fuses with other image tokens and processes through a ControlNet-like network for text generation. We visualized the attention maps of the text tokens in both U-Net and TextControlNet at three resolutions: 64x64, 32x32, and 16x16. 

In the U-Net encoder (\ding{192}-\ding{194}), the process focuses on generating image content without text, then in the TextControlNet (\ding{195}-\ding{197}), it concentrates on generating text glyphs. Finally, in the U-Net decoder (\ding{198}-\ding{200}), the integration of image and text is achieved. However, as shown in the lower part of Fig.~\ref{fig:analysis}, we discovered that TextControlNet also responds significantly to non-text tokens. This indicates that it not only facilitates text generation but also acts as part of the denoiser, working in conjunction with U-Net to generate the overall image content. While this improves the integration of image and text, it can also lead to drawbacks such as decreased overall image quality and decreased inference efficiency. This paper proposes the WriteNet+AttnX architecture to address these issues.

\section{Examples of Font Extractor and Color Picker}
\label{app: font_color}
\begin{figure*}[htbp]
 \vspace{0mm}
 \centering
 \includegraphics[width=1.0\textwidth]{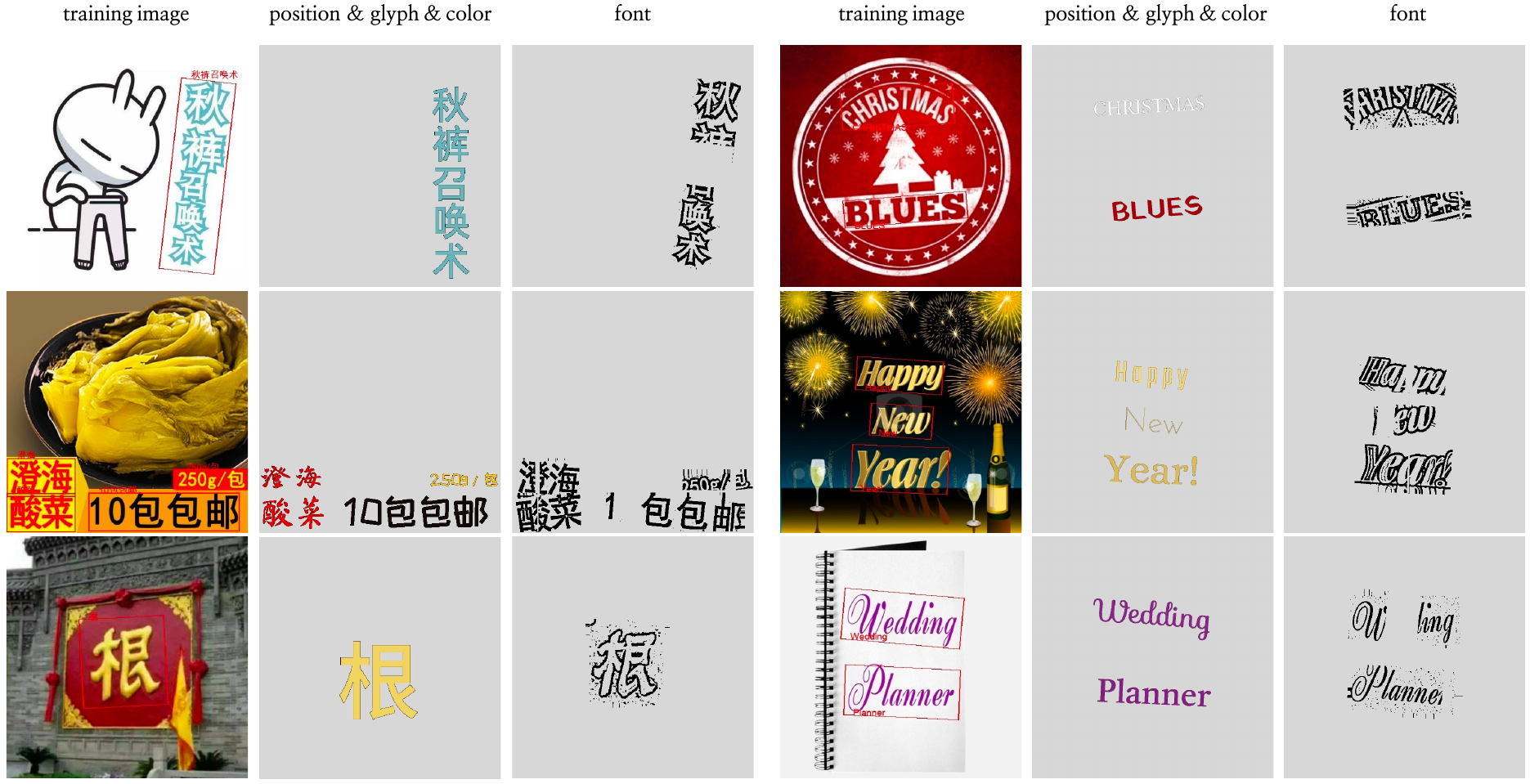}
 \vspace{0mm}
 \caption{Examples of the extracted font image and text color using the font extractor and color picker.}
 \vspace{0mm}
 \label{fig:fontcolor}
\end{figure*}

In Fig.~\ref{fig:fontcolor}, we present examples of the extracted font image $e_f$ and text color $e_c$ obtained using the font extractor and color picker. Each set contains three images: the first is the training image, the second is the glyph image $l_g$ used in the Auxiliary Latent Module, which renders each line of text onto an image according to their positions using a glyph render. For display purposes, the color $e_c$ extracted by the color picker is applied to render text. Note that during training, $l_g$ does not include color information. Moreover, each text line is rendered using a randomly selected font to prevent the leakage of font style. This also brings the advantage that AnyText2 can choose any font file to generate text during inference, unlike AnyText, which is limited to using the Arial Unicode font. The font image $e_f$ in the third image is extracted by the font extractor. To prevent the leakage of glyphs, various transformations such as rotation, translation, scaling, and occlusion are applied.

\section{Prevent Watermarks Using Trigger Words}
\label{app:watermarks}
Images containing text collected from the internet often come with numerous watermarks. According to AnyText~\cite{tuo2023anytext}, 25\% of the Chinese data and 8\% of the English data in the AnyWord-3M dataset are watermarked. They adopted a strategy of removing these watermarked images during the last two epochs of training, amounting to about 0.5 million images. We employed a different approach that, based on the watermark probability provided in AnyWord-3M, labeled as ~\textit{wm\_score}, we added ``no watermarks" to the captions with wm\_score$<$0.5, and ``with watermarks" for those with higher scores. During the inference, by adding the trigger words ``no watermarks", watermarks can be effectively prevented. The comparison with AnyText on watermark probabilities is shown in Table~\ref{table:watermark}.

\begin{table}
    \vspace{2mm}
    \small
    \centering
    \caption{\small Comparison with AnyText on watermark probabilities. }
    \vspace{0.0mm}
    \setlength{\tabcolsep}{2mm}{
    \begin{tabular}{c|c|c}
    \cline{1-3}  watermark & Chinese & English  \\ \hline
    AnyText   & 2.9\% &  0.4\% \\
    AnyText2  & 1.8\% &  0.7\% \\
    \hline
\end{tabular}}
    \vspace{-0.15in}
    \label{table:watermark}
\end{table}

\section{Parameter Size and Computational Overhead of AnyText2}
\label{app: param_size}
Our framework is implemented based on AnyText. Despite the addition of some modules, the total parameter sizes have only increased by 63.8M, as referred to in Table~\ref{table:param_size}. Moreover, due to the design of WriteNet that only performs inference once, the computational overhead is reduced. On a Tesla V100, the time taken to generate 4 images in FP16 has been reduced from 5.85s to 4.69s, resulting in a 19.8\% improvement.
\begin{table}
    \vspace{0mm}
    \small
    \centering
    \caption{\small The Comparison of the parameter sizes of modules between AnyText and AnyText2.}
    \vspace{0.0in}
    \setlength{\tabcolsep}{1.35mm}{
    \begin{tabular}{l|c|c}
    \hline
    Modules   & AnyText  & AnyText2  \\ \hline
    UNet & 859M & 859M  \\
    AttnX & - & 57M \\
    VAE & 83.7M & 83.7M \\
    CLIP Text Encoder & 123M & 123M \\
    TextControlNet/WriteNet & 360M & 360M \\
    Auxiliary Latent Module & 1.3M & 1.3M \\
    Glyph Encoder & 4.6M & 4.6M \\
    Position Encoder & - & 2.2M \\
    Font Encoder & - & 4.6M \\
    Color Encoder & - & 5K  \\ \hline
    Total & 1431.6M &  1495.4M \\
    \end{tabular}}
    \label{table:param_size}
\end{table}

\section{Examples of Long and Short Captions}
\label{app:captions}
\begin{figure*}[htbp]
 \vspace{0mm}
 \centering
 \includegraphics[width=1.0\textwidth]{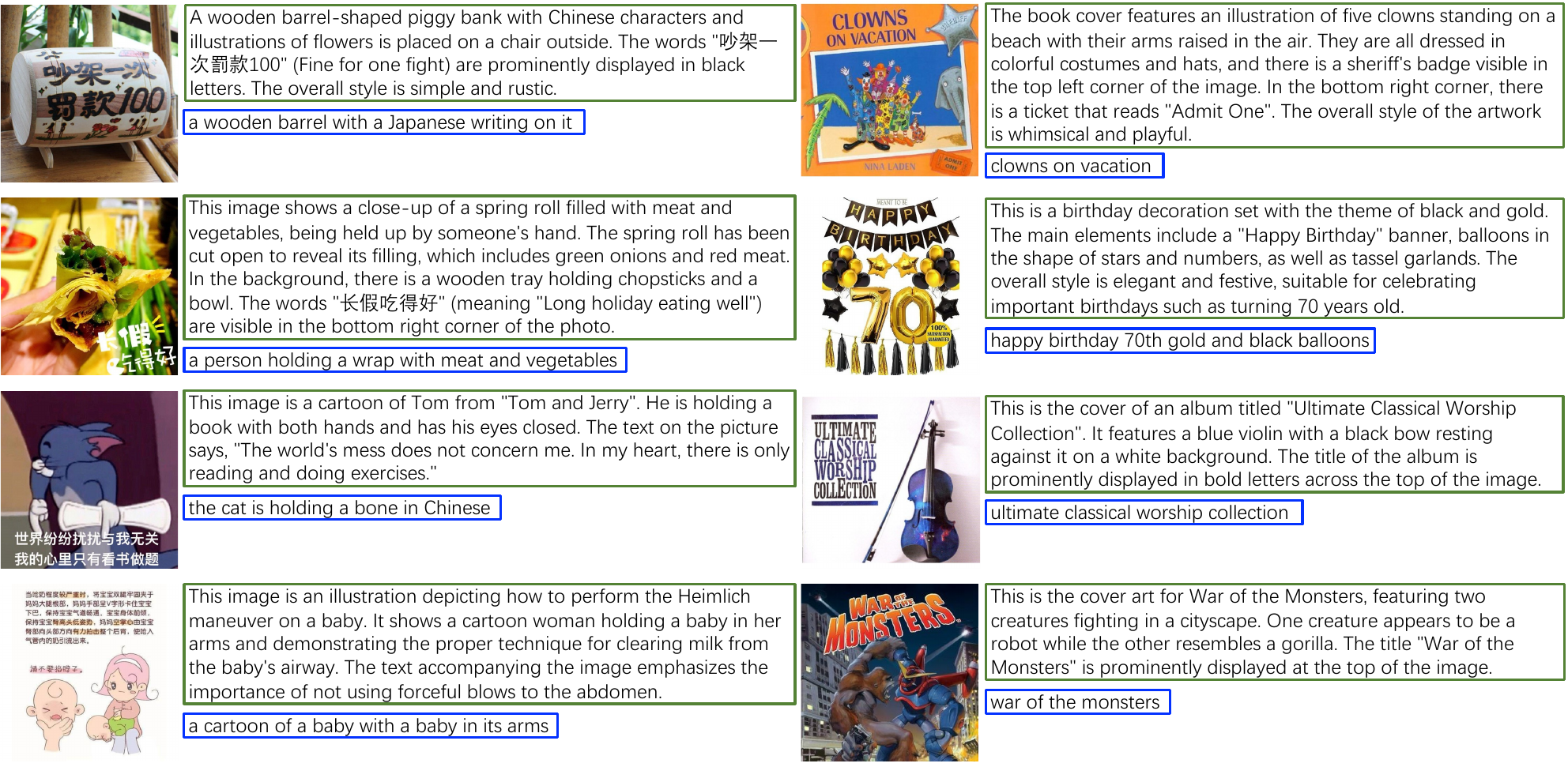}
 \vspace{0mm}
 \caption{Exmaples of training images along with long and short captions by BLIP-2 and QWen-VL.}
 \vspace{0mm}
 \label{fig:captions}
\end{figure*}

From the examples presented in Fig.~\ref{fig:captions}, it is evident that the short captions produced by BLIP-2 are very simplistic and may contain errors. In contrast, the long captions generated by QWen-VL not only provide a comprehensive description of the image details but also achieve a high level of accuracy, even accurately identifying the text within the images. We remove the quotation marks from these long captions and use them for training.

\end{document}